\documentclass[10pt,journal]{IEEEtran}
\usepackage{amsmath,amssymb}
\usepackage{booktabs,tabularx,array}
\usepackage{algorithm}
\usepackage{algorithmic}
\usepackage{url}
\usepackage[hidelinks]{hyperref}
\usepackage{graphicx}

\title{DNative-Twin: Decision Graphs and Digital Twins for Reconstructable Agentic Decisions}

\author{%
Junjie Pang\textsuperscript{1},
Zhenzhen Xie\textsuperscript{2},
Haoke Han\textsuperscript{1},
Ying He\textsuperscript{3},
Jing Wang\textsuperscript{4},
Gang Liu\textsuperscript{4,\dag}\\[4pt]
\textsuperscript{1}Qingdao University, Qingdao, China \qquad
\textsuperscript{2}Shandong University, Qingdao, China\\
\textsuperscript{3}Hangzhou Shujiao Technology Partnership (Limited Partnership), Hangzhou, China\\
\textsuperscript{4}Changchun University of Technology, Changchun, China%
\thanks{\dag\,Corresponding author: Gang Liu (\texttt{lg@ccut.edu.cn}).}%
}

\begin{document}
\maketitle

\begin{abstract}

AI agents increasingly gather evidence, invoke tools, apply constraints, and produce decisions that people or software may commit to action. A final output alone cannot show which evidence, tool state, rule, authorization, or action path produced it. We present \textbf{DNative-Twin}, a graph-native digital twin that records a committed agentic decision as a typed trajectory and re-executes its decision mechanism under declared conditions. The graph links the state observed by the agent, the path it followed, and the authority behind the resulting action. The twin synchronizes this information, replays the mechanism in isolation, and compares it under controlled changes. We instantiate the framework in enterprise decision processes using three public process logs and controlled replay suites. The experiments identify a specific failure: graph structure localizes represented changes but cannot determine the consequence of an unobserved tool state. In a three-condition controlled experiment with 300 injected instances, unresolved-divergence recall increased from 0 to 0.667 when replay-contract state was added and to 1.0 when verification results were also available; the held-out set contained no critical-class instance. Across 500--5,000 BPI 2020 cases, median end-to-end time increased from 0.794 to 8.889 seconds on the reported platform. These results separate the roles of graph structure, replay context, and verification evidence in reviewing a decision mechanism.

\end{abstract}
\begin{IEEEkeywords}AI agents, agentic decisions, decision reconstructability, digital twin, heterogeneous graph, agent governance\end{IEEEkeywords}
\section{Introduction}

AI agents now gather evidence, invoke tools, apply constraints, and recommend or execute actions in personal, organizational, and public settings. Some decisions remain under direct human review; others operate under authority delegated in advance. In both settings, a consequential action must be connected to what the system observed, how it acted, and the authority under which the action was taken.

Consider two procurement runs that both return \texttt{APPROVE}. In the first, the agent reads a current supplier alert, applies the valid purchasing policy, receives a successful budget response, and sends the recommendation to an authorized manager. In the second, the alert is stale, the budget call times out, and an informal override reaches the same label. The outputs agree, but the decision mechanisms differ. These differences determine whether the committed action can be explained, reproduced, and safely revised. We call the ability to recover this path \textbf{decision reconstructability}.

Existing operational and provenance records rarely preserve this path as one connected record. They often leave three parts disconnected: what the agent observed, how it acted, and what authorized the resulting action. Decision reconstruction requires these dependencies to be linked over time and connected to the observed outcome.

We represent this information as a \textbf{decision trajectory graph}: a dynamic heterogeneous subgraph bounded to one decision path. A real trajectory may end in commitment, while a replay may instead end in review or rejection. The graph supports reconstructability checks, constrained-path queries, and aligned comparison across real, twin, and perturbed environments. A \textbf{GraphDiff} records the differences between two aligned trajectories. The synchronized state and replay conditions provide the context needed to interpret those differences.

We implement this model in \textbf{DNative-Twin}, a graph-native digital twin for reconstructing and testing agentic decision mechanisms(Fig.~\ref{fig:architecture}). The framework first copies the relevant state into an isolated twin and replays the recorded mechanism under fixed conditions. It then changes selected conditions and compares the new trajectory with the baseline. The resulting GraphDiff shows where the mechanism changed. A rule-based review then determines whether that difference changes the decision under its declared constraints. Any proposed revision remains linked to the evidence that motivated it and must satisfy declared release conditions.

This workflow requires an identifiable decision boundary, recordable dependencies, explicit authorization, and an observable or pending outcome. Enterprise processes provide these conditions through persistent event records, explicit decision objects, and approval paths, so we first evaluate the framework in this setting. Public logs test whether the graph can be constructed from recorded events. Controlled experiments then test replay, perturbation, adjudication, and revision. In one injected suite, a graph-only detector classified all 40 unobserved tool-timeout cases as benign. In the three-condition experiment, replay-contract state recovered 60 of 90 unresolved instances, and verification results recovered all 90.

This paper makes four contributions:

\begin{enumerate}
\item We define a dynamic heterogeneous decision graph that connects what an agent observed, how it acted, and how the resulting action was authorized and committed.
\item We derive decision trajectories as decision-bounded subgraphs and formalize reconstructability for human--AI and delegated automated decisions.
\item We present DNative-Twin, which replays a decision mechanism in synchronized state and compares its trajectory under controlled changes.
\item We provide an executable specification and evaluate its enterprise instantiation on public process logs and controlled replay suites.
\end{enumerate}
Section II defines the decision graph, trajectory, and reconstructability conditions. Sections III and IV present the digital-twin architecture and its executable specification. Section V combines the enterprise case study with controlled evaluation. Sections VI--VIII discuss related work, scope, limitations, and conclusions.

\section{Agentic Decisions, Decision Graphs, and Reconstructability}

This section defines the information needed to reconstruct an agentic decision. It first represents the decision context as a dynamic heterogeneous graph. It then defines the trajectory of a committed decision, the conditions for reconstructability, and the comparison of trajectories across environments.

\subsection{Decision Graph}

We first define the state required to distinguish the two procurement runs introduced above. Let the decision-relevant state of an agentic system at time $t$ be

\begin{equation}
\mathcal{S}_t = (B_t,K_t,P_t,T_t,H_t,A_t,O_t),
\end{equation}

where $B_t$ contains decision objects; $K_t$ contains versioned knowledge; $P_t$ contains policies, constraints, and authority grants; $T_t$ contains tools and their observable states; $H_t$ contains authorized human or machine actors and their roles; $A_t$ contains agents and bounded skills; and $O_t$ contains outcome signals. $\mathcal{S}_t$ is the state relevant to one class of decisions. It includes an entity or state when changing that element can alter what an agent or authorized actor should know, do, avoid, escalate, commit, or later revise.

The decision layer is modeled as a temporal attributed heterogeneous multigraph

\begin{equation}
\mathcal{G}_t=(V_t,E_t,X_t,\phi,\psi,\rho),
\end{equation}

where $V_t$ is the set of nodes observed or valid by time $t$, $E_t \subseteq V_t\times\mathcal{T}_E\times V_t$ is the set of typed directed edges, and $X_t$ contains node and edge attributes. The maps $\phi:V_t\rightarrow\mathcal{T}_V$ and $\psi:E_t\rightarrow\mathcal{T}_E$ assign node and relation types. The metadata map $\rho$ assigns each graph element an environment label, source time, ingestion time, validity interval, version, provenance reference, and confidence when applicable. A multigraph is required because the same entities may participate in several relations or repeated time-indexed interactions.

The graph contains state, execution, authorization, and decision entities. An interaction is represented by nodes and typed edges instead of one overloaded event tuple. For example, a \texttt{ToolCall} node connects to its invoking \texttt{Agent}, input \texttt{DecisionObject}, used \texttt{EvidenceArtifact}, tool version, and produced result. A \texttt{CommitmentEvent} connects a \texttt{Recommendation} to the \texttt{AuthorizedActor}, \texttt{DecisionRole}, and \texttt{AuthorityGrant} that turn it into an action with consequences. Later \texttt{OutcomeSignal} nodes record the observed result. This factorization preserves shared dependencies: one policy can govern several actions, one evidence artifact can support several recommendations, and one authority grant can cover a bounded class of commitments.

The state tuple identifies what must be retained; the graph schema makes those dependencies queryable. Table~\ref{tab:minimum-schema} gives the smallest set of types used by the invariant checks. A domain may add or refine types while preserving the distinction between context, execution, authorization, and commitment.

\begin{table*}[t]
\caption{Minimum node and relation types in the decision graph.}
\label{tab:minimum-schema}
\centering\footnotesize
\begin{tabularx}{\textwidth}{p{0.16\textwidth}XX}
\toprule
\textbf{Category} & \textbf{Minimum node types} & \textbf{Representative edge types} \\
\midrule
Decision context & \texttt{DecisionObject}, \texttt{EvidenceArtifact}, \texttt{KnowledgeUnit}, \texttt{PolicyRule} & \texttt{about}, \texttt{supports}, \texttt{contradicts}, \texttt{version\_of}, \texttt{governs} \\
Execution & \texttt{Agent}, \texttt{Skill}, \texttt{ToolCall}, \texttt{DecisionEvent} & \texttt{performed}, \texttt{invokes}, \texttt{uses}, \texttt{acts\_on}, \texttt{produces} \\
Authorization & \texttt{AuthorizedActor}, \texttt{DecisionRole}, \texttt{ReviewEvent}, \texttt{AuthorityGrant} & \texttt{holds\_role}, \texttt{authorizes}, \texttt{commits}, \texttt{reviews}, \texttt{delegates\_to} \\
Decision & \texttt{Recommendation}, \texttt{CommitmentEvent}, \texttt{OutcomeSignal} & \texttt{recommends}, \texttt{committed\_as}, \texttt{modified\_as}, \texttt{results\_in}, \texttt{responds\_to} \\
Twin operation & \texttt{ScenarioPerturbation}, \texttt{GraphDiff}, \texttt{CandidateRevision} & \texttt{perturbs}, \texttt{replayed\_from}, \texttt{diverges\_from}, \texttt{motivates} \\
\bottomrule
\end{tabularx}
\end{table*}

Each node used in a decision trajectory records its identity, time, type, environment, and source. Additional metadata is attached only when it is needed to reconstruct a dependency: version references for evidence, execution references for tool calls, and authority references for commitments. The graph may store governed references instead of sensitive source content.

This schema connects what informed a decision, how the mechanism acted, and what authorized the commitment. The procurement case later maps these roles to purchase requests, managers, and approvals. In an automated setting, the same authority path can end at a machine actor operating within a grant issued in advance.

\subsection{Decision Trajectory and Reconstructability}

A decision mechanism is denoted by

\begin{equation}
\mathcal{M}=(Q,\pi,\Gamma,\Lambda),
\end{equation}

where $Q$ is a set of typed decision states, $\pi$ is the action policy of participating agents and actors, $\Gamma$ contains transition and path constraints, and $\Lambda$ defines commitment, escalation, and release rules. Keeping $\mathcal{M}$ separate from $\mathcal{G}_t$ allows the same mechanism to be replayed against different synchronized states and allows candidate mechanisms to be compared over the same decision object. We decompose the constraint set as

\begin{equation}
\Gamma=\Gamma_{\mathrm{path}}\cup
\Gamma_{\mathrm{temporal}}\cup
\Gamma_{\mathrm{version}}\cup
\Gamma_{\mathrm{app}},
\end{equation}

where $\Gamma_{\mathrm{app}}$ contains state-conditioned applicability constraints.

DNative-Twin distinguishes three properties that are often conflated. \emph{Traceability} establishes the source, provenance, and version of an artifact. \emph{Validity} establishes whether that artifact was legally or temporally valid when used. \emph{Contextual applicability} establishes whether a valid artifact applies to the current decision state. For a state-sensitive dependency $v$, we define

\begin{equation}
\begin{aligned}
\operatorname{App}(v,\mathcal{S}_t)
&=\mathbf{1}\!\left[\mathcal{S}_t\models A_v\right],\\
A_v
&=\operatorname{scope}(v)\\
&\quad\land\operatorname{precondition}(v).
\end{aligned}
\end{equation}

Thus, $\operatorname{Valid}(v,t)=1$ does not imply $\operatorname{App}(v,\mathcal{S}_t)=1$. A purchasing rule may remain the current valid version while applying only below a monetary threshold or outside a high-risk supplier class. The metadata map $\rho$ therefore includes machine-readable scope and precondition fields for state-sensitive policies, evidence, authority grants, skills, tools, and path rules \cite{ferraiolo2014guide}. Whenever natural-language conditions cannot be compiled into explicit predicates, the graph records the unresolved condition and routes the dependency to governed review rather than treating it as applicable by default.

For a decision object $b$, an observation interval $[t_0,t_1]$, and a terminal node $z$ associated with $b$, define the decision trajectory as

\begin{equation}
\mathcal{T}(b,z;[t_0,t_1])
=\mathcal{G}_t[V_{b,z},E_{b,z}],
\end{equation}

where $\phi(z)\in\{\texttt{DecisionEvent},\texttt{CommitmentEvent},\texttt{ReviewEvent}\}$. The set $V_{b,z}$ contains $b$, $z$, all admissible decision ancestors of $z$ in the interval, and any outcome nodes explicitly attributed to $z$. The set $E_{b,z}$ contains the typed relations among those nodes. Admissibility is defined by relations such as \texttt{uses}, \texttt{supports}, \texttt{invokes}, \texttt{produces}, \texttt{reviews}, \texttt{authorizes}, \texttt{commits}, \texttt{committed\_as}, and \texttt{results\_in}. The boundary rule excludes records that merely share a case identifier but have no decision relation to $z$.

For sequential inspection, the trajectory can be viewed as a time-ordered event sequence:

\begin{equation}
\tau_{b,z}=\operatorname{TopoTime}\!\left(
\mathcal{T}(b,z;[t_0,t_1])\right),
\end{equation}

where $\operatorname{TopoTime}$ returns a time-respecting ordering with stable tie-breaking for concurrent events. This ordered view is useful for inspection, while the graph retains shared evidence, competing recommendations, parallel tool calls, and reviews that resolve several branches.

A recommendation and a consequential commitment are distinct nodes. Let $R$ be the set of \texttt{Recommendation} nodes and $C$ the set of \texttt{CommitmentEvent} nodes. For $a\in C$, the committed trajectory $\mathcal{T}(b,a;[t_0,t_1])$ exists only when the graph contains a recommendation relation and an applicable authority path. All antecedent relations point toward the commitment event:

\begin{equation}
\begin{aligned}
&\operatorname{Committed}(\mathcal{T},a)=1\\
&\Longleftrightarrow\\[-2pt]
&\exists r\in R:\ (r,\ell,a)\in E_{b,a},\\
&\qquad \ell\in\{\texttt{committed\_as},\texttt{modified\_as}\},\\
&\exists h,g:\ (g,\texttt{authorizes},h)\in E_{b,a},\\
&\qquad (h,\texttt{commits},a)\in E_{b,a},\\
&\phi(h)=\texttt{AuthorizedActor},\quad
  \phi(g)=\texttt{AuthorityGrant},\\
&\qquad \operatorname{Valid}(g,a.time)=1,\\
&\qquad \operatorname{App}(g,\mathcal{S}_{a.time})=1.
\end{aligned}
\end{equation}

This definition records an agent recommendation and the action that gives it practical effect as separate events. In human--AI collaboration, an authorized person may accept or modify the recommendation and create a commitment event. A rejection is recorded as a \texttt{ReviewEvent} and produces no commitment event. In automated operation, an authorized machine actor may commit the output within authority granted in advance. Both forms record the source, scope, validity, and constraints of that authority.

With the graph and trajectory established, the technical problem can now be stated. For a commitment event $a$, let $\mathcal{R}_{pre}(a)=\{R_1,\ldots,R_m\}$ be the required antecedent type groups, where each $R_j\subseteq\mathcal{T}_V$ contains acceptable alternatives for one requirement. A minimal configuration contains the groups $\{\texttt{DecisionObject}\}$, $\{\texttt{EvidenceArtifact}\}$, $\{\texttt{KnowledgeUnit},\texttt{PolicyRule}\}$, $\{\texttt{Agent},\texttt{ToolCall}\}$, $\{\texttt{AuthorizedActor}\}$, and $\{\texttt{AuthorityGrant}\}$. Let $\mathcal{C}(a)$ contain the required temporal, version, applicability, cardinality, confidence, and path constraints.

Let $\operatorname{Anc}_{\mathcal{L}}(a)$ denote nodes reverse-reachable from $a$ through admissible relation types $\mathcal{L}$, and let $\operatorname{Desc}_{\mathcal{L}}(a)$ denote nodes forward-reachable through admissible outcome relations. Outcome coverage is

\begin{equation}
\begin{aligned}
&\operatorname{OutcomeCovered}(\mathcal{T},a)=1\\
&\Longleftrightarrow
\left[\exists o\in\operatorname{Desc}_{\mathcal{L}}(a):
\phi(o)=\texttt{OutcomeSignal}\right]\\
&\qquad\lor\operatorname{PendingOutcome}(\mathcal{T},a)=1.
\end{aligned}
\end{equation}

The pending state records that the outcome is expected but not yet observable. Decision reconstructability is

\begin{equation}
\begin{aligned}
&\operatorname{Recon}(\mathcal{T},a)=1\\
&\Longleftrightarrow
\left[\forall R_j\in\mathcal{R}_{pre}(a),\right.\\[-2pt]
&\qquad\left.\exists v\in\operatorname{Anc}_{\mathcal{L}}(a):
\phi(v)\in R_j\right]\\
&\qquad\land
\operatorname{OutcomeCovered}(\mathcal{T},a)=1\\
&\qquad\land
\left[\forall c\in\mathcal{C}(a):c(\mathcal{T})=1\right].
\end{aligned}
\end{equation}

Given partial observations from heterogeneous systems, the task is to construct a trajectory graph whose commitment events satisfy $\operatorname{Recon}$, preserve the separation between recommendation and commitment, and remain comparable across real, twin, and perturbed environments. This formulation distinguishes three failures. A \textbf{capture failure} occurs when a required entity or relation is absent from the graph. A \textbf{mechanism failure} occurs when a recorded trajectory violates a decision constraint. A \textbf{divergence} occurs when two otherwise valid trajectories differ across environments. The resulting record supports later analysis of correctness, fairness, and causality, each of which requires its own evidence.

Reconstructability establishes what must be present in one trajectory. Comparing the same mechanism across environments additionally requires synchronized state and aligned trajectories. The synchronization contract defines a typed mapping

\begin{equation}
\sigma:\Delta\mathcal{G}_{real}(t)
\rightarrow\Delta\mathcal{G}_{twin}(t'),
\end{equation}

where a real graph delta is normalized, validated, and projected into the twin with explicit source and ingestion times. The mapping need not preserve identity: production identifiers may be tokenized, sensitive evidence may be represented by governed references, and unavailable tools may use consequence-preserving stubs. The contract must state what is synchronized, with what freshness, under which transformations, and how failed updates are represented.

A scenario operator $W_\delta(\mathcal{G}_{twin},\mathcal{M},b)$ applies a typed perturbation $\delta$ and produces a terminal node $z^\delta$ and trajectory $\mathcal{T}_{b,z^\delta}^\delta$. The terminal node may be a \texttt{CommitmentEvent}, \texttt{ReviewEvent}, or another declared \texttt{DecisionEvent}. Terminal nodes from two environments are aligned when they share the decision object, replay lineage, mechanism version, and observation window. We write this condition as $\operatorname{Align}(z^i,z^j\mid b,\mathcal{M},[t_0,t_1])=1$; the terminal-node identifiers and types may differ. For aligned trajectories, the graph-difference operator produces

\begin{equation}
D(\mathcal{T}_{b,z^i}^i,\mathcal{T}_{b,z^j}^j)
=(\Delta V,\Delta E,\Delta X,\Delta P)
\end{equation}

and compares nodes, edges, attributes, and decision paths across environments \cite{gao2010survey}. Let $d_{ij}=D(\mathcal{T}_{b,z^i}^i,\mathcal{T}_{b,z^j}^j)$ denote the resulting \texttt{GraphDiff}. Synchronization projects observed state changes into the twin, perturbation applies a declared intervention, and $d_{ij}$ records the resulting trajectory changes. Environment labels keep perturbed data separate from real-world evidence and preserve each divergence for review.

Locating a structural change is not enough to determine its governance effect. The next definitions separate structurally different but governance-equivalent trajectories from changes that alter a required policy, authority, evidence, or outcome condition.

Two executions may retrieve independent evidence in different orders while preserving the same decision dependencies. We define governance equivalence, $\mathcal{T}_i\sim_{\mathcal{I}}\mathcal{T}_j$, when the trajectories have equivalent required-evidence coverage, applicable-policy set, authority path, critical-invariant vector, commitment class, and outcome constraints. Structural differences inside an equivalence class are \emph{benign}; differences that change one of these governance-relevant projections are \emph{material}.

For diagnosis, each change in $d_{ij}$ may receive one or more observable source labels:

\begin{equation}
\operatorname{Src}(d_{ij})\subseteq
\{\mathrm{state},\mathrm{config},\mathrm{tool},
\mathrm{stoch},\mathrm{mechanism}\}.
\end{equation}

The source labels are not mutually exclusive. State labels cover evidence, knowledge, policy, authority, or outcome changes. Configuration labels cover model, prompt, skill, seed, or decoding changes. Tool labels cover version, availability, replay mode, or returned state. Stochastic labels mark residual variation under a fixed replay contract, and mechanism labels mark an intentional revision.

Let $\mathcal{I}$ be the declared invariants and $\mathcal{P}$ the decision policy. A rule-based adjudication function

\begin{equation}
\begin{aligned}
&\operatorname{Classify}(d_{ij},\mathcal{I},\mathcal{P},\mathcal{S}_t)\\
&\quad\in
\left\{
\begin{array}{l}
\textit{benign},\ \textit{material},\\
\textit{critical},\ \textit{unresolved}
\end{array}
\right\}.
\end{aligned}
\end{equation}

returns the first applicable label in the following order. It returns \emph{unresolved} when the available evidence cannot support a governed classification. With sufficient evidence, it returns \emph{critical} for a mandatory authority, isolation, evidence, or review violation; \emph{material} for another change to a governance-relevant projection; and \emph{benign} when the trajectories remain governance-equivalent. This priority makes the four labels mutually exclusive. Frequency does not override a mandatory constraint: a less frequent trajectory may be the valid one when it satisfies a required escalation rule.

\section{DNative-Twin Architecture}
As summarized in Fig.~\ref{fig:architecture}, DNative-Twin organizes the
decision-reconstruction workflow into three stages: graph projection and
synchronization, isolated replay and perturbation, and semantic adjudication
and revision.
\subsection{Graph Projection and Synchronization}

\begin{figure*}[t]
    \centering
    \includegraphics[width=\textwidth]{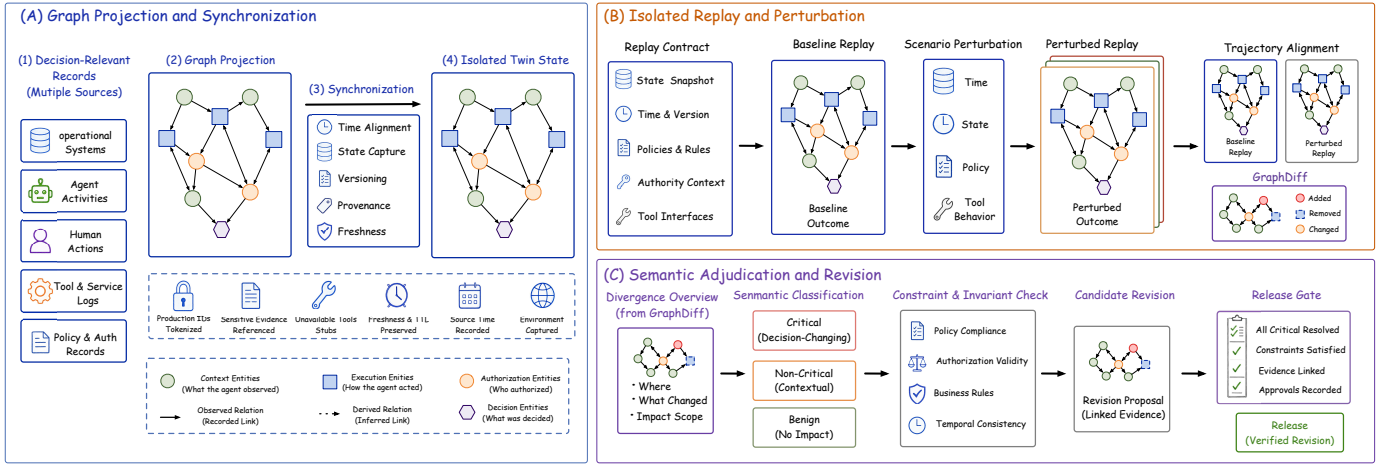}
    \caption{
Overview of DNative-Twin.
(A) Decision-relevant records are projected into a synchronized trajectory graph.
(B) The recorded mechanism is replayed in isolation under baseline and declared perturbation conditions, producing aligned trajectory differences.
(C) GraphDiffs are adjudicated against declared constraints and supporting evidence; candidate revisions are returned for verification before release, while outcome feedback updates the synchronized state.
}
    \label{fig:architecture}
\end{figure*}

Consider the purchase request used throughout the paper. When a new supplier alert arrives, DNative-Twin records the alert and its effect on the request, then copies that state into an isolated twin. The twin first replays the recorded decision and then tests a declared change, such as a budget-tool timeout. Comparing the two paths shows where the decision mechanism changed and whether the change violates a constraint. If adjustment is needed, the proposed revision is tested under the same conditions before release. Figure~\ref{fig:architecture} shows this closed loop from observation to review.

Source connectors capture decision-relevant changes from operational systems and agent activity. A normalization step converts each source record into a typed graph update while preserving its origin and time. The graph can therefore connect a decision to the information and authority on which it depended, as well as to its later outcome.

No single source record normally contains this full path. Table~\ref{tab:conceptual-coverage} therefore compares the information retained by common record types. DNative-Twin joins these records around a decision trajectory, which becomes the unit of replay and comparison.

\begin{table*}[t]
\caption{Conceptual coverage of common records. “Strong” indicates that the information is a first-class part of the record, not that every product in the category implements it.}
\label{tab:conceptual-coverage}
\centering\footnotesize
\begin{tabularx}{\textwidth}{p{0.16\textwidth}XXXX}
\toprule
\textbf{Capability} & \textbf{Workflow log} & \textbf{Agent trace} & \textbf{Provenance record} & \textbf{DNative-Twin trajectory graph} \\
\midrule
Process state and transition & Strong & Partial & Partial & Strong \\
Model/tool execution path & Weak & Strong & Partial & Strong \\
Evidence and knowledge version & Partial & Partial & Strong & Strong \\
Authorization and commitment & Partial & Weak & Partial & Strong \\
Outcome feedback & Partial & Weak & Partial & Strong \\
Synchronized shadow replay & No & Partial & No & Yes \\
Typed perturbation and cross-environment diff & No & No & No & Yes \\
\bottomrule
\end{tabularx}
\end{table*}

\subsection{Isolated Replay and Perturbation}

Isolated replay asks whether the same mechanism follows the same governed path when it runs against a synchronized copy of the decision state. For the purchase request, the twin can reuse the recorded supplier alert, policy version, and budget state without issuing another order. Any call that could affect the real environment is replaced by a read-only or simulated call labeled \texttt{environment=twin}. The twin has no authority to commit an external action.

This isolation is useful only when the real and twin runs refer to the same decision. The replay record therefore binds the real object, mechanism version, synchronization cutoff, and any controlled or substituted execution state. Comparison focuses on the resulting decision path and its governed dependencies. Thus, two runs that both approve a request still differ materially if the replay omits a required supplier alert or review. Different internal traces remain governance-equivalent when they satisfy the same declared constraints.

Under these bound conditions, shadow replay re-executes the mechanism against the synchronized state:

\begin{equation}
\begin{aligned}
\mathcal{T}_{\mathrm{shadow}}
&=\operatorname{Execute}\bigl(
\mathcal{M},\mathcal{S}^{\mathrm{twin}}_t,\\
&\hspace{5.8em}\mathcal{C}_R\bigr).
\end{aligned}
\end{equation}

The real trajectory $\mathcal{T}_{\mathrm{real}}$ is an attributable comparison reference rather than executable ground truth. The replay contract

\begin{equation}
\begin{aligned}
\mathcal{C}_R=(&b,t_{\mathrm{cut}},K,P,A,\\
               &T,H,\Theta,\Omega).
\end{aligned}
\end{equation}

groups the replay conditions into four parts: decision identity and cutoff $(b,t_{\mathrm{cut}})$; governed state $(K,P,H)$; execution configuration $(A,T,\Theta)$; and environment $\Omega$. Each unbound or substituted field is recorded as a possible source of divergence.

Tool behavior is the main case in which a replay needs a substitute. The framework supports three modes: reuse the recorded response, execute a read-only copy, or use a stub that preserves the declared result or failure. These modes answer different questions. Reusing a response isolates the downstream mechanism, whereas execution also tests the synchronized tool state. Table~\ref{tab:tool-replay-modes} states the interpretation of each mode; the \texttt{ToolCall} node records the selected mode and the information needed to reproduce it.

\begin{table}[t]
\caption{Tool-call replay modes and interpretation}
\label{tab:tool-replay-modes}
\centering
\footnotesize
\begin{tabularx}{\columnwidth}{@{}p{0.28\columnwidth}X@{}}
\toprule
\textbf{Mode} & \textbf{Question answered} \\
\midrule
Response injection & Does the downstream mechanism reproduce its governed behavior under the historical observation? \\
Executable replay & How does the mechanism behave under the currently synchronized tool state? \\
Consequence-preserving stub & How does the mechanism handle a declared result or failure without production side effects? \\
\bottomrule
\end{tabularx}
\end{table}

Replay establishes a baseline; perturbation then tests one declared dependency of that baseline. For example, the purchase-request replay can replace a successful budget check with a timeout. The resulting \texttt{ScenarioPerturbation} identifies the changed dependency and links it to the trajectory that diverges from the baseline.

Each scenario records the intervention in a machine-readable form and remains tied to the original decision and replay state. This link makes the comparison attributable: the observed difference is evaluated against a named change rather than an unspecified change in the environment.

\subsection{Semantic Adjudication and Revision}

Semantic adjudication determines whether a replay difference affects a declared constraint and whether the mechanism requires adjustment. In the budget-timeout example, a missing escalation path would be classified against the declared invariants. A candidate revision could then add the required fallback for that condition. The graph links the revision to the observed divergence and its verification record \cite{raji2020closing}. The revision becomes eligible for deployment only after it satisfies the release conditions below.

Release readiness tests whether a candidate mechanism $\mathcal{M}'$ remains within its declared constraints. The mandatory scenario suite is $\mathcal{D}_{crit}$. Its pass indicators are $C_{inv}$ for invariants and $C_{scn}$ for critical scenarios. The remaining terms measure material divergence $(r_{mat})$, governance equivalence $(r_{eq})$, twin freshness $(f_{twin})$, and unresolved critical cases $(N_{crit,unres})$ against configured thresholds. We define

\begin{equation}
\begin{aligned}
&\operatorname{ReleaseReady}(\mathcal{M}')=1\\
&\quad\Longleftrightarrow
\begin{cases}
C_{inv}=1,\quad C_{scn}=1,\\
r_{mat}\leq\epsilon_{mat},\quad r_{eq}\geq\tau_{eq},\\
f_{twin}\geq\tau_{fresh},\quad N_{crit,unres}=0.
\end{cases}
\end{aligned}
\end{equation}

Passing these criteria makes the mechanism eligible for authorization under the declared scenario suite and thresholds. The subsequent \textsc{GovApproves} step records the independent authorization decision; release requires both results. Synchronization drift, a policy or tool change, a new incident, or a new critical scenario reopens the revision cycle when it invalidates a release condition. A maximum iteration count and replay budget stop the cycle when the available evidence cannot distinguish candidate mechanisms.

Algorithm~\ref{alg:sync-replay} summarizes the graph, replay, difference, and adjudication operations defined above. Here, $S$ is the source-record collection and $\Delta_{scn}$ is the set of declared perturbations. The observation window $[t_0,t_1]$ is part of every trajectory comparison.

\begin{algorithm}[t]
\caption{Synchronize--Replay--Stress--Diff--Revise Loop}
\label{alg:sync-replay}
\begin{algorithmic}[1]
\REQUIRE $S,\sigma,\mathcal{M},b,[t_0,t_1],\Delta_{scn},\mathcal{I},\mathcal{P}$
\ENSURE Review package $R$
\STATE $G_r \gets \textsc{Observe}(S,b)$
\STATE $G_t \gets \textsc{ApplyIso}(\sigma(G_r))$
\STATE $(z_r,T_r) \gets \textsc{Trajectory}(G_r,b,[t_0,t_1])$
\STATE $(z_s,T_s) \gets \textsc{Shadow}(G_t,\mathcal{M},b,[t_0,t_1])$
\STATE $C_s \gets \textsc{Check}(T_s,\mathcal{I})$
\STATE \textbf{assert} $\textsc{Align}(z_r,z_s\mid b,\mathcal{M},[t_0,t_1])=1$
\STATE $D_s \gets \textsc{Diff}(T_r,T_s)$
\STATE $L_s \gets \textsc{Classify}(D_s,\mathcal{I},\mathcal{P},\mathcal{S}_t)$
\STATE $E \gets \emptyset$
\FORALL{$\delta \in \Delta_{scn}$}
    \STATE $G_\delta \gets \textsc{Perturb}(G_t,\delta)$
    \STATE $(z_\delta,T_\delta) \gets \textsc{Shadow}(G_\delta,\mathcal{M},b,[t_0,t_1])$
    \STATE $C_\delta \gets \textsc{Check}(T_\delta,\mathcal{I})$
    \STATE \textbf{assert} $\textsc{Align}(z_s,z_\delta\mid b,\mathcal{M},[t_0,t_1])=1$
    \STATE $D_\delta \gets \textsc{Diff}(T_s,T_\delta)$
    \STATE $L_\delta \gets \textsc{Classify}(D_\delta,\mathcal{I},\mathcal{P},\mathcal{S}_\delta)$
    \STATE $E \gets E \cup \{(\delta,C_\delta,D_\delta,L_\delta)\}$
\ENDFOR
\STATE $R \gets \textsc{Review}(C_s,D_s,L_s,E)$
\STATE $\mathcal{M}' \gets \textsc{Revise}(R)$
\STATE $q_r \gets \textsc{ReleaseReady}(\mathcal{M}')$
\STATE $q_g \gets \textsc{GovApproves}(\mathcal{M}')$
\IF{$q_r \land q_g$}
    \STATE $\textsc{Release}(\mathcal{M}')$
\ENDIF
\STATE \textbf{return} $R$
\end{algorithmic}
\end{algorithm}

\section{Checkable Properties and Minimal Artifact}

\subsection{From Architecture to Executable Specification}

The architecture becomes technically reviewable when its main claims can be checked. DNative-Twin expresses its declared decision constraints as graph properties. Table~\ref{tab:core-invariants} connects each property to a graph check and a concrete failure. Together, these checks test whether an implementation preserves the distinctions required for reconstruction and replay.

\begin{table*}[t]
\caption{Core invariants and their graph-level checks.}
\label{tab:core-invariants}
\centering\footnotesize
\begin{tabularx}{\textwidth}{p{0.16\textwidth}XX}
\toprule
\textbf{Invariant} & \textbf{Required graph check} & \textbf{Example violation} \\
\midrule
I1. Recommendation-commitment separation & \texttt{Recommendation} and \texttt{CommitmentEvent} are distinct nodes connected by a typed commitment or modification edge & Agent output is stored directly as an authorized action \\
I2. Reconstructability reachability & Every commitment event satisfies the antecedent-group and outcome-coverage checks in $\operatorname{Recon}$ & A committed action has no retrievable evidence, authority basis, or outcome linkage \\
I3. Required-review non-bypass & Every path matching a review predicate passes a valid \texttt{ReviewEvent} before commitment & A high-risk decision reaches commitment without its required review \\
I4. Twin isolation & A \texttt{CommitmentEvent} in the \texttt{twin} or \texttt{perturbed} environment has no valid \texttt{AuthorityGrant} for an external action & Shadow tool call commits a purchase order \\
I5. Knowledge-version traceability & Each trajectory dependency uses a version valid at event time and records its content reference & Decision cites “current policy” without a resolvable version \\
I6. Divergence retention & Each material real--twin or baseline--perturbed difference is stored as a \texttt{GraphDiff} linked to the aligned trajectories by typed relations & Replay differs but only the final label is retained \\
I7. Contextual applicability & Every state-sensitive dependency used by a trajectory satisfies its declared scope and preconditions under the execution state & A valid low-value purchasing rule is used for a high-value, high-risk request \\
\bottomrule
\end{tabularx}
\end{table*}

\subsection{Graph Algorithm Interface}

The invariants require six graph operations. Reverse reachability checks the dependencies required by $\operatorname{Recon}$, while constrained path queries verify review and authority paths. Temporal and applicability checks establish whether a recorded dependency was valid for the decision state. Subgraph matching detects prohibited shortcuts, such as a recommendation connected directly to an external action. Graph difference then reports changes between aligned trajectories. A property graph with a rule or query engine is sufficient to implement these operations.

\subsection{Minimal Artifact Specification}

A minimal reproducible artifact should test whether the specification can be executed outside the deployed system. Table~\ref{tab:minimal-artifact} lists the files needed for that test.

\begin{table*}[t]
\caption{Minimal Artifact Package}
\label{tab:minimal-artifact}
\centering
\footnotesize
\setlength{\tabcolsep}{3pt}
\renewcommand{\arraystretch}{1.08}
\begin{tabularx}{\textwidth}{
  @{}
  >{\raggedright\arraybackslash}p{0.36\textwidth}
  X
  @{}
}
\toprule
\textbf{Artifact} & \textbf{Purpose} \\
\midrule
\texttt{decision\_graph\_schema.json}
& Graph schema and required attributes. \\

\texttt{node\_types.json}
& Node types and type-specific fields. \\

\texttt{edge\_types.json}
& Relations and endpoint constraints. \\

\texttt{example\_real\_graph.jsonl}
& Example trajectory from the real environment. \\

\texttt{example\_twin\_graph.jsonl}
& Aligned shadow trajectory from the twin. \\

\texttt{scenario\_perturbation.json}
& Typed stress-test intervention. \\

\texttt{critical\_scenario\_suite.json}
& Mandatory scenarios used by the release check. \\

\texttt{replay\_contract.json}
& Bound state, configuration, and tool replay modes. \\

\texttt{applicability\_rules.json}
& Machine-readable scope and precondition predicates. \\

\texttt{release\_conditions.json}
& Thresholds, freshness requirement, and replay budget. \\

\texttt{check\_graph\_invariants.py}
& Invariant checks and witness diagnostics. \\

\texttt{graph\_diff.py}
& Node, edge, attribute, and path differences. \\

\texttt{classify\_divergence.py}
& Benign, material, critical, or unresolved labels. \\

\texttt{release\_readiness.py}
& Threshold checks and release-gate evidence. \\

\texttt{evaluation\_summary.json}
& Scenario results, equivalence rates, freshness evidence, and unresolved-case counts. \\

\texttt{README.md}
& Construction and execution instructions. \\
\bottomrule
\end{tabularx}
\end{table*}

The real and twin examples must share a decision object, replay lineage, mechanism version, and observation window. Each graph retains its terminal node, $z_r$ or $z_s$, so that alignment is checked before graph difference. At least one declared perturbation should then change the synchronized trajectory. The checker returns pass/fail results with witness paths or missing-type diagnostics for I1--I7, and the difference tool reports node, edge, attribute, and path changes. The scenario manifest, release conditions, and evaluation summary supply the remaining inputs to $\operatorname{ReleaseReady}$. Together, these files allow an independent implementer to construct trajectories, verify invariants, classify their differences, and execute the declared release checks without access to the deployed system.

\section{Enterprise Case Study and Controlled Evaluation}
\label{sec:enterprise-evaluation}

Enterprise processes provide persistent events, identifiable decision objects, explicit approval paths, and observable outcomes. These properties make the domain suitable for showing how the general model is instantiated and for testing its graph, replay, perturbation, and revision operations.

\subsection{Enterprise Study Design and Evidence Layers}

The enterprise evaluation separates four sources of evidence. Public logs test graph construction, while controlled overlays add decision fields that the logs do not contain. Constructed suites isolate replay and release behavior, and external-model probes measure repeated-output variation under recorded request protocols. Each field is labeled as source-observed, derived, overlaid, injected, or model-annotated.

The public-log layer uses OCEL 2.0 Procure-to-Pay \cite{park_2023_8412920, berti2024ocel}, BPI Challenge 2020 PermitLog \cite{https://doi.org/10.4121/uuid:ea03d361-a7cd-4f5e-83d8-5fbdf0362550}, and BPI Challenge 2019 \cite{https://doi.org/10.4121/uuid:d06aff4b-79f0-45e6-8ec8-e19730c248f1}. The logs record process events and object relations. LLM prompts, replay modes, authority grants, and machine-readable applicability rules appear only in the declared overlay or injected layers.

Table~\ref{tab:exp-datasets} reports the graph sizes produced by the declared mappings. The BPI 2019 result uses a seed-42 reservoir sample obtained after scanning the full log. An independent operator reran the fixed pipeline on the checksum-verified 728,558,522-byte XES and reproduced all 18 declared checks. The rerun recovered 251,734 traces, 1,595,923 events, the ordered 5,000-case sample, and the graph counts in the table; the sampled case-ID files were byte-identical.

\begin{table*}[t]
\caption{Public-log graph instantiation. Counts describe the evaluated samples.}
\label{tab:exp-datasets}
\centering
\footnotesize
\setlength{\tabcolsep}{4pt}
\begin{tabular}{@{}lrrrrrr@{}}
\toprule
Dataset & Cases/objects & Events & Nodes & Edges & Node types & Edge types \\
\midrule
OCEL 2.0 P2P & 9,543 & 14,671 & 30,956 & 81,132 & 6 & 5 \\
BPI 2020 PermitLog & 5,000 & 61,408 & 225,290 & 349,116 & 10 & 9 \\
BPI 2019 reservoir sample & 5,000 & 30,428 & 103,283 & 164,272 & 11 & 9 \\
\bottomrule
\end{tabular}
\end{table*}

The evaluation asks five questions: whether the graph can represent and retrieve decision dependencies (RQ1), whether replay is stable under a fixed contract (RQ2), whether material and null interventions can be distinguished (RQ3), whether replay-contract and verification information improve adjudication (RQ4), and whether the revision gate rejects an injected defect and accepts its bounded repair (RQ5). The comparison conditions reveal progressively more information: B0 contains only the final output, B1 adds the ordered trace, B2 adds structural GraphDiff, and B3 uses the full graph and its queries. Schema coverage reports which declared types are present; the remaining tests evaluate the behavior of the implementation.

\subsection{Illustrative Procurement Case Study}

We use a constructed procurement case to show how the enterprise instantiation operates. Purchase request \texttt{PR-2048} concerns a time-sensitive component from supplier \texttt{S-17}. The request instantiates \texttt{DecisionObject}, the procurement manager instantiates \texttt{AuthorizedActor}, and the recorded approval instantiates \texttt{CommitmentEvent}. In the baseline trajectory, procurement agent \texttt{A-proc} retrieves supplier evidence \texttt{E-supplier-v12}, applies policy \texttt{P-proc-v7}, and receives a successful response from budget tool call \texttt{TC-budget-91}. The agent produces recommendation \texttt{R-approve-1}. Manager \texttt{H-lee} reviews it under authority grant \texttt{AG-44} and creates commitment event \texttt{AD-approve-1}; the purchasing system later records outcome \texttt{O-po-issued}. This trajectory separates the recommendation from the authorized commitment and its outcome.

Table~\ref{tab:call-level-replay} contrasts the baseline record with two replay operations. A call-level execution trace can record input, output, and order. The trajectory graph also retains the dependencies needed to explain why a replay followed a different governed path.

\begin{table*}[t]
\caption{Call-level procurement replay: event recording versus executable decision-dependency analysis}
\label{tab:call-level-replay}
\centering
\footnotesize
\begin{tabularx}{\textwidth}{@{}p{0.13\textwidth}p{0.24\textwidth}p{0.25\textwidth}X@{}}
\toprule
\textbf{Stage} & \textbf{Historical log} & \textbf{Baseline replay} & \textbf{Perturbed replay and interpretation} \\
\midrule
State & \texttt{PR-2048}, policy v7, supplier evidence v12 & Same synchronized snapshot and mechanism & Supplier alert v13 or another declared state intervention \\
Tool call & \texttt{check\_budget(C18,80000)} returns \texttt{sufficient} & Response injection tests downstream reproducibility; executable replay tests synchronized tool state & Consequence-preserving stub returns \texttt{timeout} without a production write \\
Recommendation & \texttt{APPROVE} & Governance-equivalent \texttt{APPROVE} is expected under the fixed contract & \texttt{ESCALATE} is expected if the timeout or alert activates a mandatory fallback \\
Governance path & Manager review under \texttt{AG-44} & Required evidence, applicable policy, review, and authority remain reachable & GraphDiff exposes added fallback/review nodes or a missing alternate-authority path \\
Interpretation & Records what occurred & Estimates residual replay stability under controlled conditions & Tests whether relevant changes alter the governed trajectory and irrelevant changes do not \\
\bottomrule
\end{tabularx}
\end{table*}

In the perturbed scenario, a regional supplier-risk alert is added as knowledge node \texttt{K-supplier-alert-v13}. The synchronization contract copies its effective time and source reference into the twin. Policy \texttt{P-proc-v7} remains valid, but the high-risk state activates its escalation rule. Replaying the same request produces \texttt{R-escalate-2}, and the GraphDiff links the changed supplier state to the newly activated path.

The scenario engine next removes the primary manager's authority. This change exposes a structural defect: the mechanism requires review but has no valid alternate path, so the request remains suspended. The resulting review package links the stalled trajectory to the missing delegation. A candidate revision adds a bounded alternate-approver rule and is replayed against the full scenario suite before release. Two additional scenarios test evidence availability and regional shipping constraints.

The case connects the formal definitions to one complete procurement trajectory. Synchronization introduces a declared knowledge change, shadow execution avoids another purchase order, and perturbation tests a failed condition. The GraphDiff then distinguishes a valid escalation from a missing authority path. The quantitative tests below evaluate the corresponding operations.

\subsection{Controlled Evaluation Results}

\textbf{Graph construction and dependency queries.}

Table~\ref{tab:exp-datasets} shows that the implementation instantiated both case-centric and object-centric logs. On OCEL, all 3,371 evaluable commitments passed recommendation--commitment separation and reconstructability reachability, and all 1,598 purchase-order objects contained a Create--Approve--Pay trajectory. Checks that required an absent field were evaluated only after the field was supplied by the declared mapping layer.

Six mappings specified before the held-out evaluation were applied to the same 5,000 BPI 2020 cases. The base mapping produced 225,290 nodes; strict and broad commitment mappings produced 188,093 and 228,761. Removing the overlay policy layer reduced knowledge-policy traceability from 1.0 to 0, while unresolved resource identity reduced the human-approval-path pass rate from 1.0 to 0. The strict-evidence variant equaled the base mapping on this sample. The query results therefore depend on preserving the entities and relations required by each question.

Seven controlled cases answered five dependency questions. Overall query accuracy was 0.06 for B0, 0.14 for B1, 0.20 for B2, and 1.00 for B3. B3 executed graph queries over task-specific cases and overlay ground truth; B0--B2 lacked part of the information available to B3. This experiment shows that the graph queries recover the declared dependencies in the seven constructed cases.

A separate annotation probe tested whether event-log text alone supplied stable governance labels. On 20 BPI 2020 cases, two prompt-separated LLM annotators achieved 0.55 raw agreement ($\kappa=0.5175$).

\textbf{Replay, perturbation, and adjudication.}

Fifty real BPI 2020 case carriers were executed 30 times under a fixed temperature-zero contract. Exact trajectory match and governance-equivalent replay rate were 1.0, and no trajectory variation was observed in this configuration. A separate injected-noise condition produced a flip rate of $0.098\pm0.014$ while preserving governance equivalence. This condition tests whether governance equivalence remains stable when outputs are deliberately changed; external-model variation is measured separately below. Separately, 38 of the 50 selected carriers passed I5, which measures knowledge-version traceability.

The controlled perturbation suite applied 42 material and 27 null interventions to seven constructed cases. Ground truth used an applicability threshold of 50,000, while the detector used an alert band of 55,000 specified before the evaluation. It detected 36/42 material changes (PS=0.8571; F1=0.9231), and all 27 null changes remained stable (NPS=1.0). The six misses were near-threshold changes from 50,000 to 50,001, which identifies the detector's boundary sensitivity.

We then injected seven drift types into 40 held-out real BPI 2020 carriers, yielding 280 instances. Sensitivity over material and critical interventions, benign stability, and source localization were each 1.0. A graph-only detector classified all 40 unresolved tool-timeout instances as benign, giving zero unresolved recall and macro-F1 0.6667. An unrepresented tool state leaves no structural evidence from which to determine its consequence.

A three-condition controlled experiment specified before the held-out evaluation isolated the required information. Thirty held-out BPI 2020 carriers were combined with 11 declared scenarios. The critical-timeout scenario required an adopted request above 50,000; all 30 carriers were at or below this threshold, so those instances were excluded. The remaining 300 instances comprised 180 benign, 30 material, and 90 unresolved cases, with no critical case. Condition A used graph signals, Condition B added replay-contract tool state, and Condition C additionally used verification-result fields.

\begin{table}[t]
\caption{Semantic adjudication under three information conditions in the controlled injected protocol.}
\label{tab:exp-e7}
\centering
\footnotesize
\setlength{\tabcolsep}{3pt}
\begin{tabular}{@{}lccc@{}}
\toprule
Metric & A: Graph & B: +Contract & C: +Verification \\
\midrule
Macro-F1 & 0.600 & 0.908 & 1.000 \\
Unresolved recall & 0.000 & 0.667 & 1.000 \\
Unresolved$\rightarrow$benign FN & 1.000 & 0.333 & 0.000 \\
Critical support & 0 & 0 & 0 \\
\bottomrule
\end{tabular}
\end{table}

As Table~\ref{tab:exp-e7} shows, replay-contract state recovered 60 of the 90 injected unresolved instances, and verification-result fields recovered all 90. Condition C uses injected signature-verification results and therefore measures the value of making those outcomes available to adjudication. The reported macro-F1 is averaged over the supported benign, material, and unresolved classes.

The three tool modes clarify why the contract matters. With no drift, response injection, executable replay, and consequence-preserving stubbing agreed. Under budget or supplier drift, injection retained the historical observation, executable replay exposed the current state change, and the stub preserved its decision consequence without a production write. Tool replay mode is therefore part of the meaning of a trajectory comparison.

\textbf{Revision, runtime, and external-model probes.}

The revision demonstration introduced $\mathcal{M}_0$, in which a budget timeout had no fallback, and $\mathcal{M}_1$, which added bounded escalation. Critical-invariant pass rate increased from 0.7143 to 1.0, material-divergence rate decreased from 0.3333 to 0, governance-equivalent rate increased from 0.7143 to 1.0, and unresolved critical cases decreased from six to zero. Under the four evaluated thresholds, $\mathcal{M}_0$ failed and $\mathcal{M}_1$ passed. Applying the gate to a release additionally requires synchronization freshness, the complete critical-scenario suite, iteration limits, and replay-budget checks.

Scalability was measured at four BPI 2020 sample sizes with five repetitions and no failed run. Median end-to-end time was 0.794, 1.682, 4.335, and 8.889 seconds at 500, 1,000, 2,500, and 5,000 cases, respectively. At 5,000 cases, median graph construction and invariant checking required 1.797 and 3.846 seconds. These measurements report observed scaling for Python 3.11.15 and NetworkX 3.6.1 \cite{SciPyProceedings_11} on an Apple M5 Pro with 64 GB RAM.

The external-model probes used five repeated calls per case. Baseline agreement is the fraction of repeated outputs that matched the recorded baseline label. The exploratory 20-case DeepSeek probe and the preregistered 100-case probes produced mostly parseable outputs, with exact-repeat rates from 0.70 to 0.96. Table~\ref{tab:exp-llm-noise} reports the full results.

\begin{table}[t]
\caption{External-model noise probes on real BPI 2020 case carriers.}
\label{tab:exp-llm-noise}
\centering
\footnotesize
\setlength{\tabcolsep}{3pt}
\begin{tabular}{@{}lrrr@{}}
\toprule
Probe & Valid calls & Exact repeat & Baseline agree. \\
\midrule
DeepSeek-20 & 100/100 & 0.700 & 0.820 \\
DeepSeek-100 & 500/500 & 0.870 & 0.932 \\
Kimi-100 & 496/500 & 0.960 & 0.982 \\
\bottomrule
\end{tabular}
\end{table}

The 100-case probes used the same carrier set and prompt but different models and provider-imposed sampling parameters. The probes therefore characterize repeated-output variation within each recorded protocol. Four Kimi responses were unparseable, and its interrupted run resumed from per-case checkpoints recorded in metadata. The remaining variation motivates repeated replay when output stability is part of the contract.

The evaluation supports three findings. The declared mappings construct typed graphs from the reported enterprise logs. The controlled suites execute replay, perturbation, adjudication, and bounded constraint adjustment under their specified conditions. In the injected protocols, graph structure misses an unrepresented tool failure, while replay-contract state and injected verification results recover the 90 unresolved instances in the three-condition experiment.

\section{Related Work}

Agent systems interleave model inference with external actions, making tool behavior and environmental state part of the safety boundary. ReAct established a reasoning--acting pattern in which language models alternate between reasoning and actions against external sources \cite{Yao2023ReAct}. ToolEmu introduced an LM-emulated sandbox for testing high-stakes tools and long-tailed failure scenarios without invoking production services \cite{Ruan2023ToolEmu}. Agent-SafetyBench broadens this empirical line with interactive environments and tool-mediated safety cases \cite{Zhang2024AgentSafetyBench}. At the organizational level, the NIST Generative AI Profile recommends continuous monitoring, provenance-aware documentation, and human intervention processes across the AI lifecycle \cite{NIST2024GAIProfile}.

Sandbox and guardrail work tests whether an action can proceed safely. Decision reconstruction adds a record-level question: which observed state and authorization condition produced the action, and how does that path change under replay? DNative-Twin addresses this question by retaining the governed trajectory and its cross-environment difference.

Provenance systems describe how entities, activities, and agents contribute to the production or transformation of an artifact. The W3C PROV data model provides interoperable concepts and validity constraints for such lineage records \cite{W3C2013PROVDM}. Distributed tracing provides a complementary operational view: OpenTelemetry represents execution as a graph of spans with identifiers, timestamps, attributes, events, status, and causal links \cite{OpenTelemetry2026Trace}. Recent agent-specific work enriches this substrate. AgentTrace organizes runtime observations into cognitive, operational, and contextual surfaces and connects structured agent logging to telemetry infrastructure \cite{AlSayyad2026AgentTrace}.

For decision reconstruction, this lineage continues through the authorized actor who commits or modifies a recommendation and the authority grant that permits the action. The resulting record also supports synchronized replay and comparison, rather than serving only as a historical trace.

Business process management separates process design, execution, monitoring, and improvement. Process mining connects these activities by deriving behavioral models from event data and checking observed execution against them \cite{vanderAalst2012Manifesto}. Agentic BPM extends this setting to agents that observe process state and act through organizational tools. Proposed architectures cover the data, reasoning, action, and orchestration needed for such systems \cite{Dumas2026ABPMS}, while the Agentic BPM manifesto emphasizes bounded autonomy and explainability \cite{Calvanese2026ABPMManifesto}.

Building on process-mining foundations, Park and van der Aalst describe a digital twin of an organization as a process-oriented representation connected to operational data and action-oriented analysis \cite{Park2021DTO}. Organization-design research distinguishes digital models, shadows, and twins by the direction and degree of coupling with organizational reality \cite{Becker2023DTO}. For agentic decisions, the synchronized projection must retain the state that affected the governed path. DNative-Twin uses isolated execution to test that path before a revised mechanism is released.

Knowledge graphs represent typed entities and relations, while temporal knowledge graphs associate facts with time points or validity intervals. Surveys of knowledge-graph representation and temporal completion describe methods for link prediction, temporal reasoning, extrapolation, and dynamic representation learning \cite{Ji2022KGSurvey,Cai2022TKGSurvey}. These methods provide useful machinery for evolving multi-relational enterprise data, especially when facts, identities, and dependencies change over time.

Here, typed schema and explicit graph queries turn temporal relations into inspectable checks for missing dependencies, invalid authority, and cross-environment differences. Process models describe flow, guardrails constrain actions, traces record execution, and provenance records lineage. DNative-Twin links these records through a committed decision and makes the trajectory executable under a replay contract.

\section{Discussion and Limitations}

DNative-Twin applies when an agentic process has an identifiable decision object and a bounded trajectory whose dependencies, constraints, authority, and outcome can be recorded. These conditions make reconstruction and replay well defined. They can occur in enterprise, personal, clinical, and public-sector settings, under either human review or delegated automation. The framework targets consequential processes that cross a decision boundary; generation without a committed action falls outside this scope.

The instantiation evaluated in this paper uses enterprise processes because their persistent events, approval paths, and outcomes make the required conditions observable. Public logs test graph construction and structural queries. Overlays supply decision fields absent from those logs, and controlled injections isolate selected mechanism failures. Evaluation in another domain requires evidence, authority rules, constraints, and outcomes defined for that domain.

The experiments also show why representation, replay, and verification are separate layers. A graph can localize a represented change, but an unobserved tool state leaves no structural evidence of its consequence. The replay contract records the tool state and expected behavior, while verification results record whether the relevant condition held. These sources complement the graph when the decision depends on information outside the recorded trajectory.

The evidence is bounded by its representation and evaluation settings. Schema design and event-to-graph extraction determine what can be reconstructed, and some queries depend on derived or overlay fields. The BPI 2019 measurements describe the reported sample. Replay stability was measured under one fixed configuration, and the external probes cover two provider-specific models under recorded sampling conditions. The adjudication evidence covers injected benign, material, and unresolved cases, with no held-out critical instance.

The release demonstration evaluates four of the eight declared conditions. A deployment must also evaluate freshness, the complete critical-scenario suite, iteration limits, and replay budget. Because the graph records sensitive operational relationships, deployment also requires access control, retention limits, and privacy-preserving references.

\section{Conclusion}

Reconstructing an agentic decision requires the mechanism that produced and authorized it, not only its final output. DNative-Twin records that mechanism as a trajectory graph and re-executes it under declared conditions. In the enterprise instantiation, the reported mappings construct decision graphs from case-centric and object-centric public logs together with declared overlay fields. The controlled experiments then exercise replay, perturbation, adjudication, and bounded constraint adjustment. In the three-condition injected experiment, unresolved recall increased from 0 to 0.667 with replay-contract state and to 1.0 when injected verification results were available; the held-out set contained no critical instance. On the reported platform, 5,000 BPI 2020 cases were processed in a median 8.889 seconds. The resulting trajectory differences show which state, path, or authorization condition changed before a revised mechanism is released.

\bibliographystyle{IEEEtran}
\bibliography{DNative-Twin_refs_v0.9}
\end{document}